\documentclass{article}
\usepackage{todonotes}
\usepackage[final,nonatbib]{acv}
\usepackage[utf8]{inputenc} 
\usepackage[T1]{fontenc}    
\usepackage{url}            
\usepackage{booktabs}       
\usepackage{amsfonts}       
\usepackage{nicefrac}       
\usepackage{microtype} 
\usepackage{multirow}

\title{DAD vision: opto-electronic co-designed computer vision with division adjoint method}

\author{
  Zihan Zang \\
  Tsinghua University\\
  \texttt{zangzh17@mails.tsinghua.edu.cn} \\
  \And 
  Haoqiang Wang \\
  Tsinghua University\\
  \texttt{whq18@mails.tsinghua.edu.cn} \\
  \And 
  Yunpeng Xu \\
  Tsinghua University\\
  \texttt{xyp21@mails.tsinghua.edu.cn} \\
}

\begin{document}
\maketitle

\begin{abstract}
 The miniaturization and mobility of computer vision systems are limited by the heavy computational burden and the size of optical lenses. Here, we propose to use a ultra-thin diffractive optical element to implement passive optical convolution. A division adjoint opto-electronic co-design method is also proposed. In our simulation experiments, the first few convolutional layers of the neural network can be replaced by optical convolution in a classification task on the CIFAR-10 dataset with no power consumption, while similar performance can be obtained.
\end{abstract}

\section{Introduction}

With the rise of deep-learning, a large number of computer vision tasks have been explored, resulting in a series of new products, such as face recognition devices and robots. In recent years, with the development of the mobile Internet and the deployment of high-speed 5G networks, a large number of new Internet-of-thing (IoT) devices and wearable devices are emerging, such as AR/VR glasses and so on. These devices have put forward higher requirements on the size and energy efficiency of computer vision systems.

\begin{figure}[h]
\caption{Typical computer vision system and proposed opto-electronic co-designed system}
\includegraphics[width=0.8\textwidth]{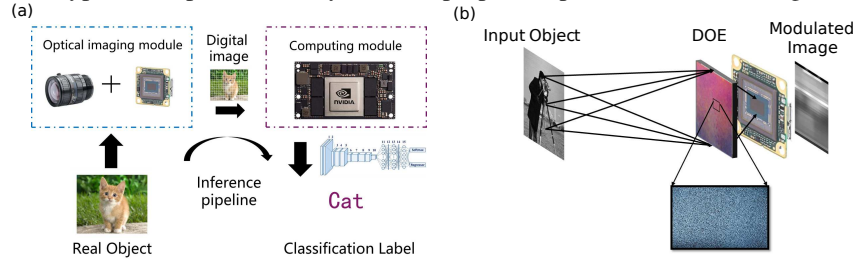}
\label{fig:vis}
\centering
\end{figure}

The inference process of a typical computer vision system is shown in Fig. \ref{fig:vis}(a). It consists of two parts: the optical imaging module and the computing module. In the image classification task, the optical module captures a clear image of a real object, and the computing module feeds the image into a neural network to obtain a classification label for the captured object. The optical system is often composed of bulky multiple lenses to correct the optical aberration, which prevents further miniaturization of the computer vision system. As for computing module, in order to realize higher classification accuracy, the number of parameters and the amount of operations in the neural network are very large, which brings great pressure on the selection of computing hardware and the power budget of the system. Some approaches, such as model pruning \cite{Liu2017learning} and quantization training \cite{Quantization} techniques, are proposed to reduce the number of parameters and operations of the neural network while the accuracy of the network is maintained.

However, there is another hidden computing capability with the potential to reduce the computation complexity of electronics part, which is the optical module. Most optical elements, such as lens, diffusers and phase plates are passive elements, which are inherently energy efficient and could modulate the light field in speed of light . The use of optical elements in analog computing has been intensively studied for decades \cite{oc}. Recently, optical elements has been proposed to implement optical neural network \cite{2017np,2018Science,PRX_optical}. However, all-optical computing suffers from substantial restrictions such as low accuracy, lack of nonlinear operations and limited reconfigurability. Therefore, given these limitations, it is wise to allocate computing tasks to both optical elements and electronics. When optical systems is involved in the computation pipeline of computer vision tasks such as image classification, it has the potential to improve the power efficiency of the system. Based on this idea, we proposed the opto-electronic co-designed computer vision system as shown in Fig. \ref{fig:vis}(b). A single ultra-thin diffractive optical elements (DOE) is used instead of conventional imaging lenses. In contrast, this passive device cannot form a clear image on the image sensor but acts as first several layers of neural networks by directly modulating the input light field, thus reducing the computation complexity and energy consumption of the back-end neural network. Also, benefited by the compact single-element DOE, the volume of the system can be reduced, leading to ultra-compact lensless computer vision system.

In order to realize the above concept, it is necessary to explore operations that can be well implemented via optical elements, and then discuss the co-design method for both the optical components and the back-end neural network under specific scenarios. According to Fourier optics, an optical system can usually be regarded as a linear invariant system (LIS). The output of an optical system can be calculated by the convolution of the input and the impulse response, which is also known as the point spread function (PSF). By using this PSF model, the most straightforward co-design approach is substituting the optical system with an one channel convolution layer, whose convolution kernel is a trainable PSF, but there are two main problems . First, the PSF is impulse response of the optical system, which is the intensity distribution pattern generated by an ideal point source after passing through the optical system, thus the intensity values are non-negative, and the sum of all intensity values are 1 in total. Therefore, the optical system can only represent a convolution operation with a non-negative convolution kernel, which has limited expressiveness and requires additional constraints on the convolution kernel during training. Second, since the convolution achieved by the optical system has the advantage of zero energy consumption, the size of a convolution kernel should be created as large as possible to maximize the advantage. However, the design of large convolutional kernels is rarely seen in deep neural networks, and the training of large convolution kernels has greater difficulties.

\begin{figure}[h]
\caption{A diagram of our proposed division adjoint (DAD) method.}
\includegraphics[width=0.8\textwidth]{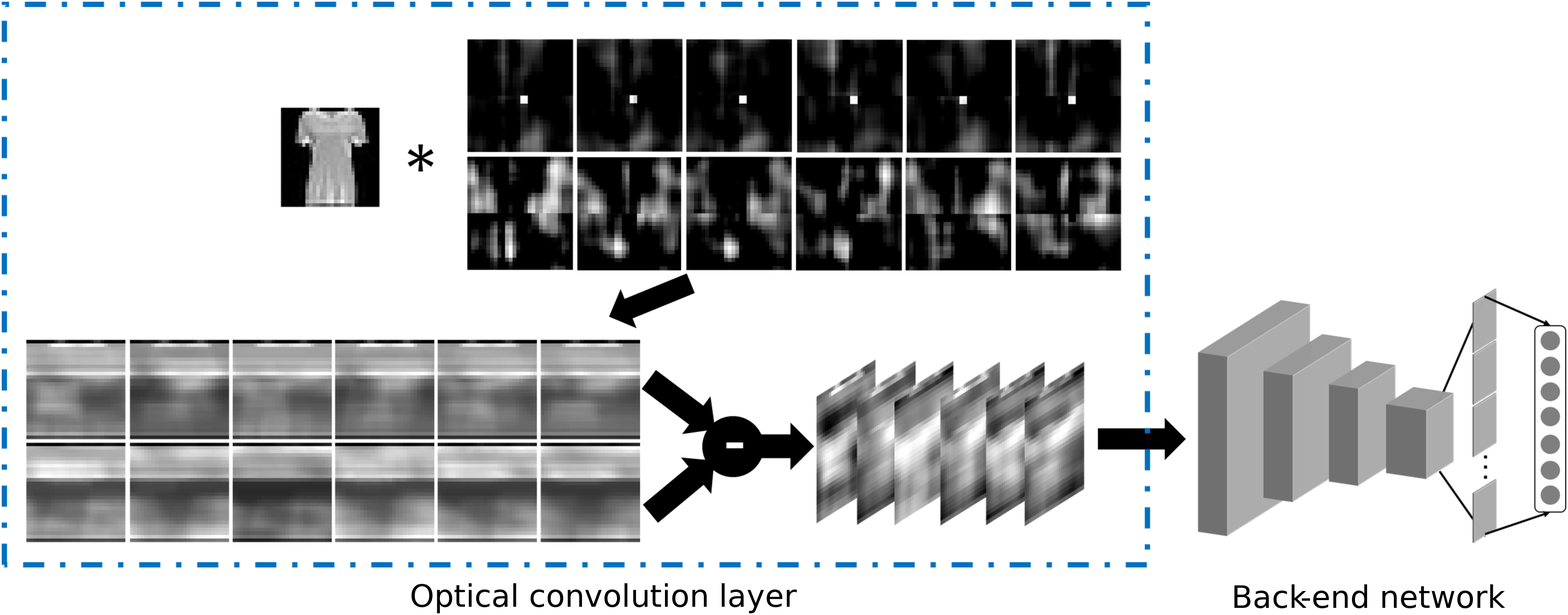}
\label{fig:alg}
\centering
\end{figure}

To address these two problems of the vanilla co-design method, we propose a division adjoint (DAD) method, which is shown in Fig.\ref{fig:alg}. The idea is mainly about dividing the computation results of optical convolutional into small pieces by designed division approach, and the differences between the small ``pieces'' are fed into the back-end networks by multiple channels. F. Yonggan \textit{et al.} \cite{Codesign} adopted similar method for adjoint design, but no detailed description and physical experiments were performed in their work.

On CIFAR-10 classification task, our proposed opto-electronic co-design architecture with DAD method reduces up to $\mathbf{33\%}$ multiply–accumulate (MAC) operations and achieves the comparable top-1 accuracy compared to the baseline with electronic only neural networks.

\section{Related works}
\subsection{Lensless imaging and computational imaging by PSF engineering}
Based on the PSF model mentioned above, lensless imaging and computational imaging is enabled by PSF engineering. J. Wu \textit{et al.} \cite{fzadnn} utilizes Fresnel zone aperture like PSF and builds a deep neural network based image reconstruction method to decode the modulated image on CMOS sensor. J. Chang \textit{et al.} \cite{Depth} utilizes the distance-dependence property of PSF and employs neural networks to recover the depth information, leading to single-shot and single-lens depth imaging. In order to solve the problem of high dynamic range imaging with single exposure, C. A. Metzle \textit{et al.} \cite{HDR} uses a sparsely distributed PSF to alleviate detector saturation. With a specially designed neural network, high dynamic range images can be generated with only one shot.

\subsection{Computational holography and the design of diffractive optical elements (DOEs)}
In order to form the desired PSF at the detector, the incoming light is phase-modulated by an very thin optical elements. As for the forward precess, Fourier optics\cite{FO} theory can precisely predict the intensity distribution on the focal plane for any given phase modulation. For the inverse problem, various methods have been proposed to calculate the phase distribution that can form any given pattern at a certain distance after the modulator. This inverse problem is often referred to as phase retrieval, which is also the key to holographic display. Gerchberg–Saxton algorithm \cite{GS} is the most popular phase retrieval method, while the quality of the reconstructed image is severely limited. Recently, gradient descent for the problem has been enabled by Writinger flow \cite{WF}, which shows much improved reconstruction quality.

\section{Proposed approach}
\begin{figure}[h]
    \caption{Explanation of division adjoint (DAD) method.}
    \includegraphics[width=0.8\textwidth]{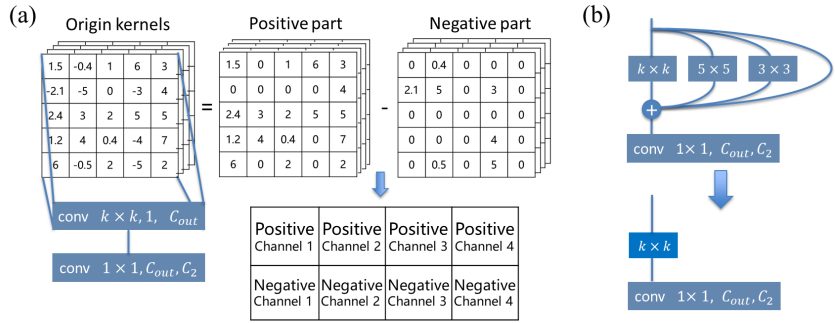}
    \label{fig:opticl_conv}
    \centering
 \end{figure}

 Considering the linearity of the convolution operations, a multi-channel convolution kernels could be considered as the difference of two positive-elements only kernels, which is shown in Fig. \ref{fig:opticl_conv}(a). In order to implement these two multi-channel positive-elements only convolution kernels with one channel optical convolution, kernels of each channel are stitched together to generated a large kernel. Therefore, using our division adjoint (DAD) method mentioned above, optical convolution could be mathematically modeled as a multi-channel convolution layer and trained together with the back-end neural networks. 
 
 Compared with the modern CNNs, the output channels of the DAD method are still not enough, which is always less than 12 and limited by the total size of the PSF. An $1\times 1$ convolution layer is utilized after optical convolution layer for increasing output channels. In our co-design architecture, optical convolution layer is used to replaced the first several convolution layers in electronic-only architecture, which is always a stack of the $3\times 3$ convolution layers. In order to obtain the same expression capability using optical convolution with only two convolution layers, first, SiLU active function is adopted to provide enhanced nonlinear performance. Second, larger kernel size $k$ is chosen to make the parameters of optical layer are comparable with the replaced convolution stack. As shown in Fig. \ref{fig:opticl_conv}(b), multi-branch structure with smaller convolution kernel size such as $7\times 7$, $5\times 5$ and $3\times 3$ is adopted during training pipeline to improve performance of the optical layer. After training, re-parameterization \cite{repvgg} technique is used to remove auxiliary branch, which padding small kernels with zeros and sum the auxiliary branch to the desire large kernel branch.
 
 After the design of PSFs, a stochastic-gradient-descent (SGD)-based phase retrieval algorithm is used to generate the phase modulation applied by the DOE. In real fabrication process of DOEs, the phase modulation is achieved by direct writing lithography. To further compensate for the fabrication errors, the lithography fabrication model of the DOE is also inserted in the phase retrieval algorithm, leading to an end-to-end  algorithm. This algorithm can directly generate dose distribution of lithography from the target PSFs.

\section{Experiments}
\begin{figure}[h]
    \caption{Neural network structure and co-design pipeline}
    \includegraphics[width=1.0\textwidth]{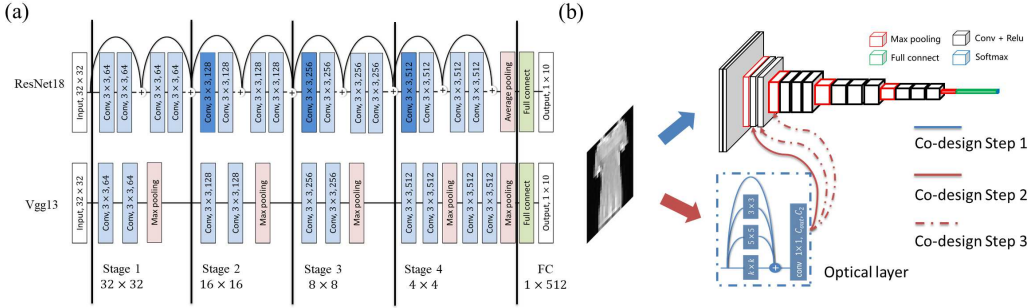}
    \label{fig:co-design}
    \centering
 \end{figure}

In our experiments, CIFAR-10 dataset is chosen to validate the performance of our co-design method, which consists of 60000 $32\times 32$ colour images in 10 classes. There are 50000 and 10000 images for training and test. We employ two widely used CNN architectures, VGG13 and ResNet18, with our DAD co-design method, the profile of this two architectures is shown in Fig. \ref{fig:co-design}(a). The co-design pipelines is shown in Fig. \ref{fig:co-design}(b), using VGG architecture as an example, firstly, we training the standard structure and find the top-1 accuracy. Secondly, we replaced the first stage of the VGG with the optical layer mentioned in section 3 and then using the same training settings with the first step to training this co-designd structure. Thirdly, we further improve the kernel size and remove the convolutions layers in stage two of the VGG. After several attempts, largest reduction of multiply–accumulate operations while keeping the same top-1 accuracy could be obtained. 
 
As for experiments and training settings, basic data arguments strategies (e.g. random crop and flip) and basic color normalization are employed. We set batch size to 256 and train models on one NVIDIA RTX A4000(16GB) GPU. SGD optimizer is used and we set the momentum to 0.9 and weight decay to $5e^{-4}$. Training epochs is set as 100 and the learning rate is start from 0.1 and reduces using a cosine annealing schedule. In experiments, multiply–accumulate (MAC) operations of neural nertowks model is counted by THOP \cite{thop} library. 

\begin{figure}[h]
    \caption{PSF pattern used in experiments.}
    \includegraphics[width=0.8\textwidth]{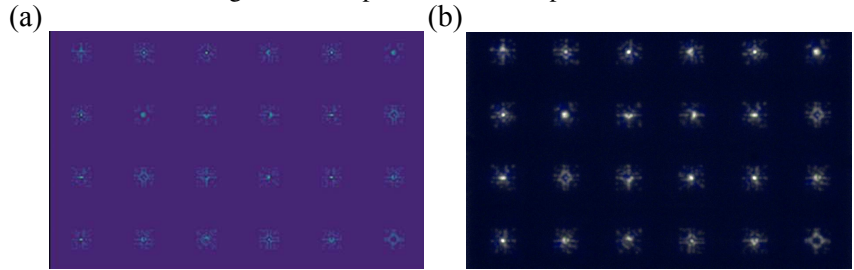}
    \label{fig:psf}
    \centering
 \end{figure}

The final experiments results is shown in Table. \ref{experiments}, in experiments of VGG13, co-design method could achieve $91.86\%$ top-1 accuracy, which is $0.05\%$ less than the electronic only model, the MAC operations of co-design model is reduced about $33.2\%$. As for ResNet18, co-design method could achieve $93.36\%$ top-1 accuracy and the MAC operations of co-design model is reduced about $30.7\%$. The optical layer of two experiments are all set to 12 channel and the kernel size k is 13. After re-parameterization, the final PSF is shown in Fig .\ref{fig:psf}(a), the real PSF of fabricated DOE captured by CMOS sensor is shown in Fig .\ref{fig:psf}(b).

\begin{table}[t]
\caption{Experiments results}
 \label{experiments}
 \centering
\begin{tabular}{l|cc|cc}
\toprule
\multirow{2}{*}{models} & \multicolumn{2}{c}{VGG13}            & \multicolumn{2}{c}{ResNet18}         \\
                        & Electronic Only & Co-designed method & Electronic Only & Co-designed method \\
\midrule
Top-1 Acc:              & 91.91\%         & 91.86\%            & 93.02\%         & 93.36\%            \\
MAC (MFLOPs)            & 229             & 153                & 488             & 338                \\
MAC reduction           &                 &\textbf{33.2\%}     &                 & \textbf{30.7\%}             \\
\bottomrule
\end{tabular}
\end{table}

\bibliographystyle{ieee.bst}
\bibliography{acv.bib}

\begin{thebibliography}{10}\itemsep=-1pt

\bibitem{thop}
Thop: Pytorch-opcounter.
\newblock \url{https://github.com/Lyken17/pytorch-OpCounter}.

\bibitem{WF}
E.~J. Candes, X.~Li, and M.~Soltanolkotabi.
\newblock Phase retrieval via wirtinger flow: Theory and algorithms.
\newblock {\em IEEE Transactions on Information Theory}, 61(4):1985–2007, Apr
  2015.

\bibitem{Depth}
J.~Chang and G.~Wetzstein.
\newblock {Deep optics for monocular depth estimation and 3D object detection}.
\newblock {\em Proceedings of the IEEE International Conference on Computer
  Vision}, 2019-Octob:10192--10201, 2019.

\bibitem{repvgg}
X.~Ding, X.~Zhang, N.~Ma, J.~Han, G.~Ding, and J.~Sun.
\newblock Repvgg: Making vgg-style convnets great again.
\newblock In {\em Proceedings of the IEEE/CVF Conference on Computer Vision and
  Pattern Recognition}, pages 13733--13742, 2021.

\bibitem{FO}
O.~K. Ersoy.
\newblock {\em Diffraction, fourier optics, and imaging}.
\newblock Wiley, 2007.

\bibitem{Codesign}
Y.~Fu, Y.~Zhang, Y.~Wang, Z.~Lu, V.~Boominathan, A.~Veeraraghavan, and Y.~Lin.
\newblock {SACoD: Sensor Algorithm Co-Design Towards Efficient CNN-Powered
  Intelligent PhlatCam}.
\newblock {\em Proceedings of the IEEE/CVF International Conference on Computer
  Vision (ICCV)}, (2):5168--5177, 2021.

\bibitem{PRX_optical}
R.~Hamerly, L.~Bernstein, A.~Sludds, M.~Solja{\v{c}}i{\'{c}}, and D.~Englund.
\newblock {Large-Scale Optical Neural Networks Based on Photoelectric
  Multiplication}.
\newblock {\em Physical Review X}, 9(2), may 2019.

\bibitem{Quantization}
B.~Jacob, S.~Kligys, B.~Chen, M.~Zhu, M.~Tang, A.~Howard, H.~Adam, and
  D.~Kalenichenko.
\newblock Quantization and training of neural networks for efficient
  integer-arithmetic-only inference.
\newblock In {\em Proceedings of the IEEE Conference on Computer Vision and
  Pattern Recognition (CVPR)}, June 2018.

\bibitem{2018Science}
X.~Lin, Y.~Rivenson, N.~T. Yardimci, M.~Veli, Y.~Luo, M.~Jarrahi, and A.~Ozcan.
\newblock {All-optical machine learning using diffractive deep neural
  networks}.
\newblock {\em Science}, 361(6406):1004--1008, sep 2018.

\bibitem{Liu2017learning}
Z.~Liu, J.~Li, Z.~Shen, G.~Huang, S.~Yan, and C.~Zhang.
\newblock Learning efficient convolutional networks through network slimming.
\newblock In {\em ICCV}, 2017.

\bibitem{HDR}
C.~A. Metzler, H.~Ikoma, Y.~Peng, and G.~Wetzstein.
\newblock {Deep Optics for Single-Shot High-Dynamic-Range Imaging}.
\newblock {\em Proceedings of the IEEE Computer Society Conference on Computer
  Vision and Pattern Recognition}, pages 1372--1382, 2020.

\bibitem{2017np}
Y.~Shen, N.~C. Harris, S.~Skirlo, M.~Prabhu, T.~Baehr-Jones, M.~Hochberg,
  X.~Sun, S.~Zhao, H.~Larochelle, D.~Englund, and M.~Soljacic.
\newblock {Deep learning with coherent nanophotonic circuits}.
\newblock {\em Nature Photonics}, 11(7):441--446, 2017.

\bibitem{oc}
G.~W. Stroke.
\newblock Optical computing.
\newblock {\em IEEE Spectrum}, 9(12):24--41, 1972.

\bibitem{fzadnn}
J.~Wu, L.~Cao, and G.~Barbastathis.
\newblock Dnn-fza camera: a deep learning approach toward broadband fza
  lensless imaging.
\newblock {\em Opt. Lett.}, 46(1):130--133, Jan 2021.

\bibitem{GS}
G.-z. Yang, B.-z. Dong, B.-y. Gu, J.-y. Zhuang, and O.~K. Ersoy.
\newblock Gerchberg–saxton and yang–gu algorithms for phase retrieval in a
  nonunitary transform system: a comparison.
\newblock {\em Applied Optics}, 33(2):209, Jan 1994.

\end{thebibliography}

\end{document}